\documentclass{article}
\usepackage{iclr2027_conference,times}
\usepackage[T1]{fontenc}
\usepackage{amsmath,amsfonts,amssymb}
\usepackage{graphicx}
\usepackage[table]{xcolor}
\usepackage{url}
\usepackage{hyperref}
\usepackage{xspace}
\usepackage{bm}
\usepackage{booktabs}
\usepackage{multirow}
\usepackage{adjustbox}
\usepackage{pifont}
\usepackage[capitalize]{cleveref}

\newcommand{\dataset}{PanoScan\xspace}

\definecolor{scenecolor}{rgb}{0.95,0.95,0.75}

\Crefname{section}{Section}{Sections}
\crefname{section}{Sec.}{Secs.}
\Crefname{equation}{Equation}{Equations}
\crefname{equation}{Eq.}{Eqs.}
\Crefname{figure}{Figure}{Figures}
\crefname{figure}{Fig.}{Figs.}
\Crefname{table}{Table}{Tables}
\crefname{table}{Tab.}{Tabs.}


\title{E2Pano: Learning Event-to-Panorama Image Reconstruction}
\author{
Zhenyang Li$^{\dagger}$ \quad
Zongqi He$^{\dagger}$ \quad
Jia Pan$^{*}$ \quad
Shijie Lin$^{*}$ \quad
Yifan Peng$^{*}$ \\
The University of Hong Kong \\
Hong Kong, China \\
{\normalfont\small $^{\dagger}$Equal contribution. $^{*}$Corresponding authors.}
}

\iclrfinalcopy

\begin{document}
\maketitle
\lhead{}
\begin{figure}[h!]
    \centering
    \includegraphics[width=\linewidth]{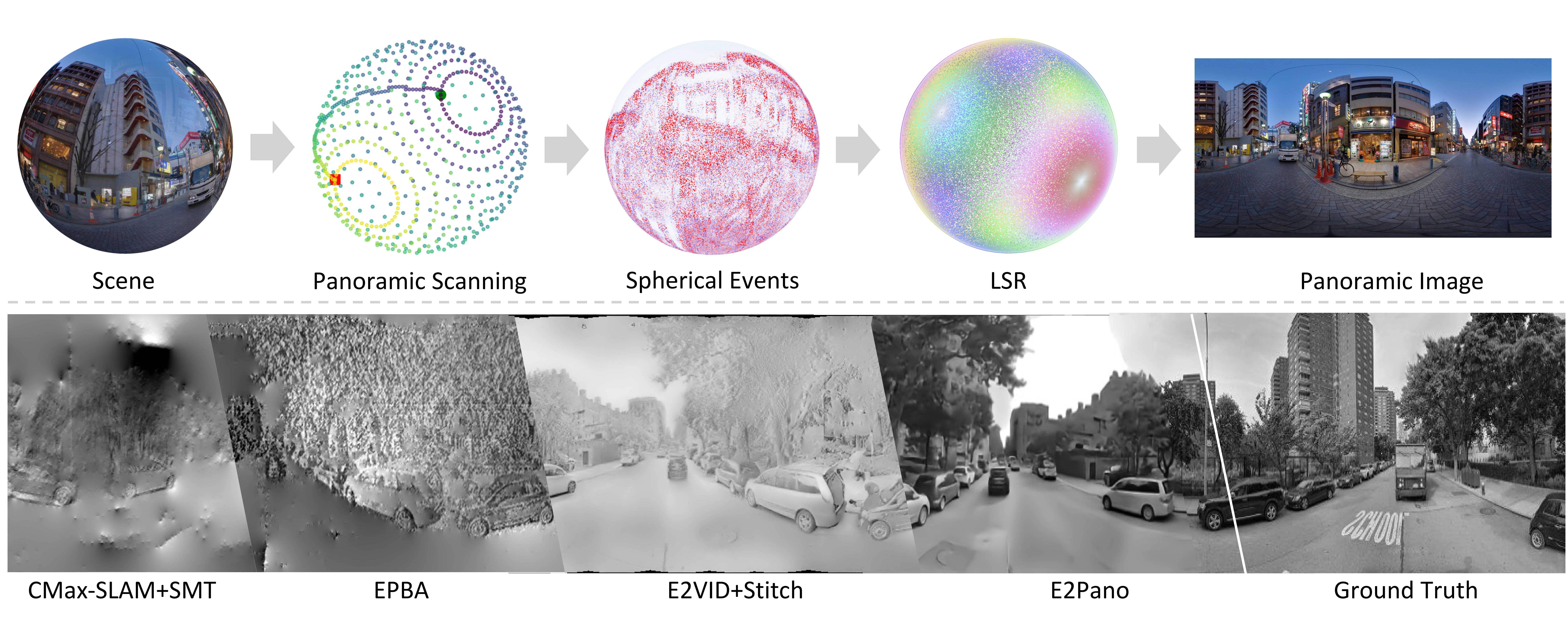}
    \caption{\textbf{Event-to-Panorama (E2Pano) overview.}
    \emph{Top}: Our pipeline uses spherical projection to transform raw event
    data into geometrically consistent event point clouds, which are then
    reconstructed into photometric panoramic images
    ($8{,}631\times2{,}000$ resolution) with an end-to-end learnable spherical
    reconstruction stage (LSR). \emph{Bottom}: Compared with optimization-based
    methods such as CMax-SLAM~\citep{guo2024cmax}+SMT~\citep{smt},
    Event-based Photometric BA (EPBA~\citep{guo2025event}), and
    E2VID~\citep{rebecq2019events}+Stitch~\citep{brown2007automatic}, our
    method yields sharper and more coherent reconstructions on rotational
    scans.}
    \label{fig:teaser}
\end{figure}

\begin{abstract}
    Event cameras offer microsecond-level temporal resolution and high dynamic range, potentially facilitating motion-blur-free panoramic imaging from fast rotational scanning. Nonetheless, existing optimization-based methods remain computationally demanding, while prior learning-based reconstruction methods are largely designed for perspective imagery and lack geometry-aware support for panoramic outputs.
    We present E2Pano, a geometry-guided event-to-panorama pipeline with an end-to-end learnable photometric reconstruction stage. Our framework preserves real spherical coordinates from geometric mapping throughout the pipeline, employs a lightweight enhancement module with frequency-domain supervision to bridge the event-image domain gap, and leverages a spherical Transformer with 3D positional embeddings for photometric reconstruction.
    Experiments on synthetic data and captured rotational scans show improved reconstruction quality and lower photometric reconstruction cost than optimization-based baselines, together with encouraging transfer to real captures under our acquisition protocol despite training purely on synthetic data.
    Additionally, we construct PanoScan, a dataset with 4,370 synthetic and 30 real-world panoramic scenes paired with event streams. Our dataset and code will be released.
\end{abstract}

\section{Introduction} 
\label{sec:intro}

Panoramic imaging facilitates wide field-of-view (FoV) photography, e.g., up to 360$^\circ$$\times$180$^\circ$, supporting various applications, including virtual reality~\citep{kim2016vr360}, scene understanding~\citep{schonberger2018semantic}, and computational photography~\citep{shum2000systems,brown2007automatic}.
Conventional approaches to capturing panoramas, such as camera scanning or multi-camera rigs~\citep{eden2006seamless}, often struggle with two problems: \textit{limited dynamic range}, which leads to degraded performance in high-contrast scenes (e.g., night-time or indoor-outdoor transitions); and \textit{slow acquisition speed}, necessitating extensive frame overlap~\citep{uyttendaele2001eliminating}.
These concerns can significantly hinder deployment in challenging real-world environments such as HDR indoor-outdoor inspection, low-light urban panoramic surveying, and pre-captured VR/telepresence content acquisition.
In contrast, event cameras~\citep{lichtsteiner2008dvs,brandli2014240}, as bio-inspired sensors, encode brightness changes as asynchronous event streams~\citep{gallego2020event}.
They offer distinct advantages for panoramic imaging: their high dynamic range (up to 140~dB) eliminates exposure bracketing in high-contrast scenes, microsecond-level temporal resolution avoids motion blur during fast rotational sweeps, and asynchronous sensing enables efficient data capture across wide FoV without frame synchronization overhead, making them a compelling alternative to conventional frame-based systems.

Nevertheless, reconstructing high-fidelity panoramas from event streams remains challenging, primarily due to scalability constraints.
The task requires handling ultra-high-resolution, sparse spatiotemporal signals, e.g., hundreds of millions of asynchronous events, to recover photometric content. 
Classical optimization-based pipelines, including contrast maximization~\citep{guo2024cmax} and photometric bundle adjustment~\citep{guo2025event}, process events sequentially, resulting in high computational cost that limits scalability for long rotational scans and large outdoor scenes.
Yet, existing learning-based reconstruction methods such as E2VID~\citep{rebecq2019events} and HyperE2VID~\citep{scheerlinck2020fast}, while effective for perspective intensity recovery, are tailored to limited-FoV planar projections and often suffer from severe distortions when applied to spherical panoramas.
More critically, these architectures are not geometry-aware: they assume perspective planar imaging and lack inductive biases for spherical/equirectangular projections, making them brittle under non-Euclidean geometry, varying intrinsics, and diverse scanning trajectories. 
Consequently, they remain difficult to deploy in panoramic settings with notable FoV changes, rapid rotations, and depth-induced parallax. Overall, prior approaches do not yet provide a geometry-aware learning pipeline for panoramic intensity reconstruction under rotational scanning.

In light of these, we present a \emph{geometry-guided Event-to-Panorama (E2Pano) pipeline} with an end-to-end learnable photometric reconstruction stage for high-fidelity panorama reconstruction from raw events.
We first perform spherical event mapping that projects asynchronous events onto the unit sphere, and employ ES-ICP~\citep{xing2023eroam} to estimate a geometrically consistent spherical representation that preserves true 3D viewing directions and scanning poses. 
Building on this geometry, we introduce a lightweight residual event enhancement module together with a learned spherical representation (LSR), trained with a magnitude-and-phase frequency loss, to bridge the event-to-panorama gap, transforming sparse event accumulations into encoder-compatible, feature-rich inputs. 
Finally, a spherical-aware Transformer decoder with 3D spherical positional embeddings reconstructs photometric panoramas. 
This design improves \emph{scalability} for photometric reconstruction on millions of events and supports transfer across scenes, optics, and rotational scan patterns. Throughout this paper, we focus on camera-center-fixed, rotation-dominated panoramic scanning rather than general free-motion event reconstruction. We summarize our main contributions as follows:
\begin{itemize}
    \item We present a geometry-guided event-to-panorama (E2Pano) image-processing pipeline that combines a geometric front-end with an end-to-end learnable photometric reconstruction stage.
    \item We introduce a lightweight event enhancement module and a learned spherical reconstruction scheme with 3D spherical positional embeddings and complementary spectral supervision.
    \item We construct \dataset, a benchmark with 4,370 synthetic scenes and 30 real-world rotational captures covering $360^{\circ}\times50^{\circ}$ FoV, and demonstrate effective reconstruction on both synthetic and captured data. Dataset and code will be released.
\end{itemize}

\begin{figure}[t]
    \centering
    \includegraphics[width=\linewidth]{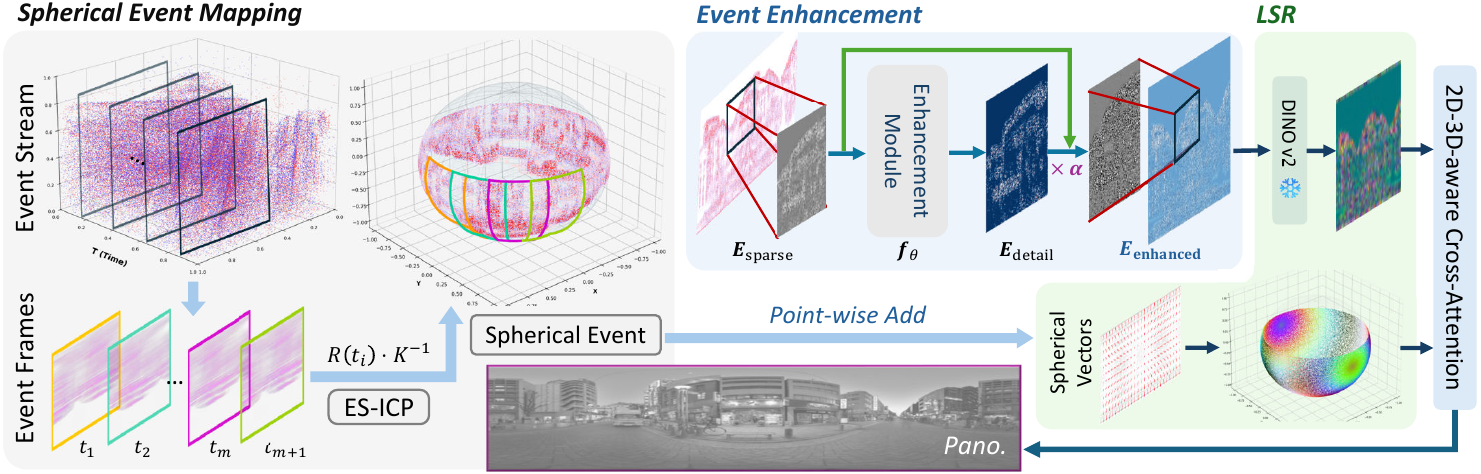}
    \vspace{-17pt}
    \caption{
    \textbf{E2Pano Framework Overview.} Our two-stage pipeline transforms raw event streams into high-fidelity panoramic images through three stages: (1) Spherical Event Mapping projects events onto a unit sphere via ES-ICP~\citep{xing2023eroam}, producing an equirectangular representation and preserved 3D coordinates; (2) Event Enhancement employs a lightweight residual network trained with frequency-domain loss to bridge the event-image domain gap; (3) Spherical Reconstruction uses a geometry-aware Vision Transformer with 3D positional embeddings derived from preserved spherical coordinates to reconstruct photometric panoramas in an end-to-end learnable photometric stage.
    }
    \label{fig:pipeline}
    \vspace{-6pt}
\end{figure}

\section{Related Work} 
\label{sec:related}

\subsection{Learning-based Image Stitching}
Image stitching serves as a fundamental tool for panorama generation, aligning and compositing overlapping views into seamless wide-FoV outputs. Deep learning reshapes this task from traditional multi-stage pipelines to end-to-end trainable frameworks. Early unsupervised methods~\citep{nie2021unsupervised} adopt two-stage architectures (coarse alignment + reconstruction) optimized via perceptual and SSIM losses, avoiding reliance on ground-truth labels. 
To handle non-projective distortions from camera translation, hybrid deformation models~\citep{nie2022deep} combine global homography with thin-plate spline warping for local non-rigid alignment, preserving foreground shapes while smoothing backgrounds. 
Recent architectures explore progressive refinement~\citep{lin2023stitched} and hybrid deformation models with learned geometric priors~\citep{nie2022deep}. 
However, these methods target narrow-FoV RGB image pairs with substantial overlap, relying on dense photometric correspondences, and thereby struggle with event cameras' sparse, asynchronous data and wide-FoV panoramic scenarios lacking dense texture, limiting applicability to our task. 

\subsection{Event-based Image Reconstruction}
Event cameras~\citep{lichtsteiner2008dvs,brandli2014240,gallego2020event} asynchronously report per-pixel brightness changes with microsecond temporal resolution and exceptional dynamic range. Event-based SLAM methods~\citep{rebecq2017ems,zhou2021event,vidal2018ultimate} demonstrate strong robustness in geometric tracking but typically output sparse point clouds without photometric intensity reconstruction. 
To recover intensities, event-to-image reconstruction evolves from hand-crafted filtering and variational methods~\citep{bardow2016simultaneous} to learning-based recurrent models~\citep{rebecq2019events,scheerlinck2020fast,wang2020event,stoffregen2020reducing}, and more recently to transformers, diffusion models, and neural radiance fields~\citep{weng2023event,zhang2023evdiff,rudnev2023evnerf}. Prior work such as Stoffregen et al.~\citep{stoffregen2020reducing} further highlights the importance of event-threshold diversity for sim-to-real transfer. 
Noteworthy, these methods are largely tailored to perspective, narrow-FoV settings and mostly operate as standalone photometric modules, neither leveraging rotational panoramic geometry nor explicitly addressing equirectangular reconstruction.

\subsection{Event-based Panoramic Reconstruction}
Relevant works primarily focus on rotational panoramic reconstruction through non-learning optimization paradigms. Event Generation Model (EGM)-based methods explicitly model the event triggering process. For instance, SMT~\citep{smt} introduces a dual-filter architecture combining particle filtering for tracking with pixel-wise EKFs for gradient mapping, while RTPT~\citep{rtpt} formulates pose estimation as energy minimization over a probabilistic event occurrence map. These methods are closely related to our task because they directly target rotational panoramic mapping, but they remain sensitive to noise and quantization while becoming costly at high resolution.

Contrast Maximization (CM)-based methods~\citep{cm-w,gallego2018unifying} estimate motion by maximizing the sharpness of warped event images. CM-GAE~\citep{cm-gae} reduces drift through global event alignment, while CMax-SLAM~\citep{guo2024cmax} develops a complete rotational SLAM system whose panoramic map is obtained as a by-product. These methods provide strong geometry baselines for rotational scans, yet they rely on iterative warping and optimization over long event sequences, which makes large-scale photometric reconstruction expensive and sensitive to local optima~\citep{shiba2022event}.

Photometric optimization under rotational motion is studied by EPBA~\citep{guo2025event}, which is the closest baseline to our work because it jointly refines geometry and photometry from event streams. Compared with alternatives like SMT and CMax-SLAM, EPBA reconstructs intensity more explicitly; compared with our method, however, it remains an optimization-based formulation whose photometric stage scales with iterative refinement over the event sequence. Our work instead couples a geometry front-end with an end-to-end learnable photometric back-end operating on spherical event maps.

\section{Method: E2Pano Pipeline} 
\label{sec:pipeline}

The goal is to transform raw asynchronous event streams obtained from continuous rotational scanning into high-fidelity, wide-FoV panoramic images. To achieve this, our E2Pano framework tackles two key challenges. First, the method must reduce the cost of photometric reconstruction for long event sequences without relying entirely on scene-specific iterative optimization. Second, it must preserve the spherical geometry induced by camera-center-fixed rotational scanning rather than approximating it with planar projections. As shown in Fig.~\ref{fig:pipeline}, we address these challenges through a hybrid two-stage pipeline: a geometric stage that aggregates events on a spherical manifold, followed by an end-to-end learnable photometric stage that enhances event maps and reconstructs panorama intensities while respecting the preserved geometry.

\subsection{Spherical Event Mapping}
\label{sec:spherical_mapping}

Panoramic event capturing from rotational scanning can generate hundreds of millions of asynchronous events, each encoding a single pixel brightness change at microsecond precision. Any practical pipeline must still read these events once, so our goal is not to avoid the linear event-loading cost itself. Instead, we aim to reduce the cost of downstream iterative photometric optimization over the raw event stream. Sequential methods~\citep{guo2024cmax,guo2025event} repeatedly operate on events in temporal order, which makes long scans expensive. To reduce this burden and enable parallel reconstruction in the learned back-end, we project events onto a structured 2D spherical manifold, transforming the temporal stream into a spatial representation amenable to neural processing.

Given raw events $\mathcal{E} = \{e_i: (x_i, y_i, t_i, p_i)\}$, where $(x_i, y_i)$ is the pixel location, $t_i$ is the timestamp, and $p_i \in \{-1, +1\}$ is the polarity, we accumulate them onto the sphere through pose-aware projection. We employ spherical Iterative Closest Point (ICP) to estimate camera rotations $\mathbf{R}(t) \in SO(3)$ by matching accumulated event point clouds to a global spherical map. The camera intrinsics and lens distortion are calibrated offline, and the camera motion is modeled as pure rotation around a fixed optical center. Detailed hardware specifications, calibration parameters, and effective field-of-view measurements are provided in the Supplementary Material. Each event is then undistorted using camera intrinsics $\mathbf{K}$ and projected to the unit sphere:
\begin{equation}
    \mathbf{V}_i = \mathbf{R}(t_i) \mathbf{K}^{-1} [ x_i, y_i, 1 ]^\text{T}, 
    \quad \mathbf{V}_i \leftarrow \frac{\mathbf{V}_i}{\|\mathbf{V}_i\|},
\end{equation}
where $\mathbf{V}_i \in \mathbb{R}^3$ represents the 3D unit vector on the sphere. Events are accumulated into a spherical point cloud $\mathcal{P}_{\text{sphere}} = \{(\mathbf{V}_i, c_i)\}$, where intensity $c_i$ aggregates polarities $p_i$ within local neighborhoods.

To handle large-scale scans, we employ regional density control by dividing the sphere into angular bins and dynamically downsampling over-represented regions while preserving details in sparse regions~\citep{xing2023eroam}. This strategy ensures memory efficiency without sacrificing coverage.

The computational role of this stage is also important. ES-ICP remains an optimization-based geometric front-end whose cost depends on local correspondence search and the number of iterations required for convergence. However, once events are aggregated into a fixed-resolution spherical map, the learned photometric stage operates on the projected representation rather than repeatedly revisiting the full event stream. This separates the unavoidable geometric alignment cost from the downstream reconstruction cost, whose complexity is governed primarily by the spherical map resolution and decoder token count.

The final output is an equirectangular projection of size $H \times W$, where each pixel $(u,v)$ stores two channels of information: an intensity channel $I(u, v)$ encoding the accumulated event count, and a geometry channel $\mathbf{V}(u, v) \in \mathbb{R}^3$ preserving the 3D unit vector from the mapping stage.
Critically, we preserve the exact 3D coordinates $\mathbf{V}$ computed during pose estimation, rather than recomputing them from pixel indices. Similar spherical direction parameterizations have been used in panoramic vision, but here $\mathbf{V}$ is taken directly from the ES-ICP/EROAM mapping output and carried into reconstruction, which differentiates our pipeline from methods that synthesize coordinates only from the equirectangular image grid. This design maintains geometric consistency between mapping and reconstruction (justified in Sec.~\ref{sec:reconstruction}). The dependence on ES-ICP also defines an explicit limitation: if geometric alignment fails, the subsequent photometric reconstruction inherits the error and may fail accordingly.

\subsection{Event Enhancement Module}
\label{sec:enhancement}

\begin{figure}[t]
    \centering
    \includegraphics[width=0.75\linewidth]{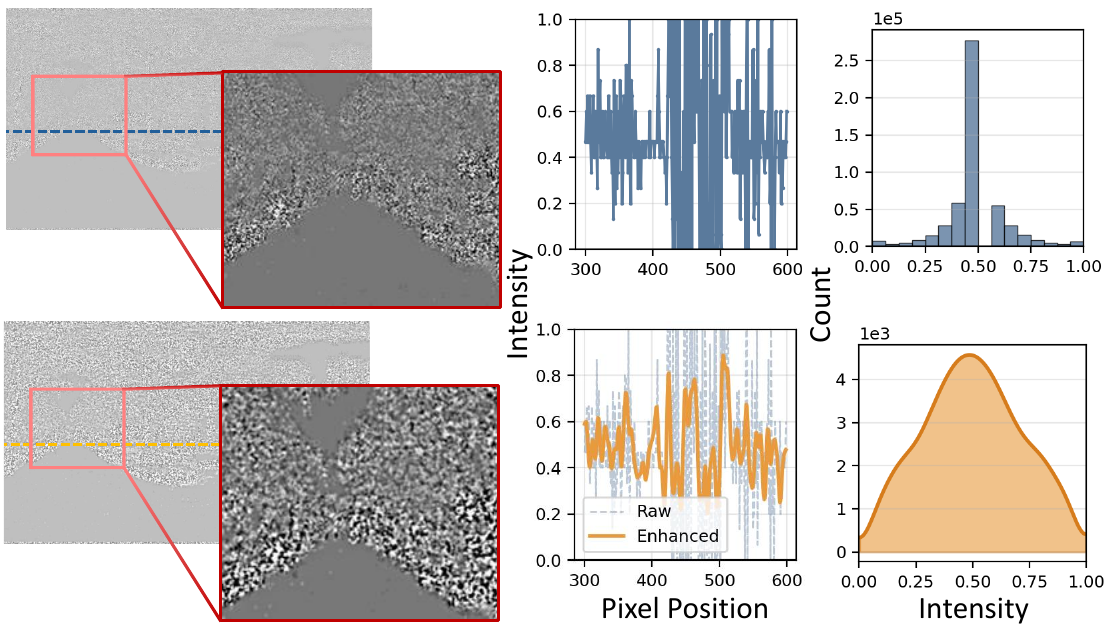}
    \vspace{-15pt}
    \caption{\textbf{Event-Image Domain Gap.} Comparison of raw (\emph{top}) v.s. enhanced (\emph{bottom}) events. Columns: (\emph{left}) accumulated images with marked scanline and zoom-ins to red squares, (\emph{center}) pixel intensity profiles showing stepwise quantization (blue, 16 levels) v.s. smooth gradients (orange), (\emph{right}) histograms demonstrating sparse clustering v.s. continuous distribution. Our enhancement bridges the domain gap from sparse and discrete events to natural continuous image statistics.}
    \label{fig:domain_gap}
    \vspace{-9pt}
\end{figure}

A fundamental challenge in learning-based event processing is the domain gap between accumulated event representations and natural images. When events are aggregated into 2D maps, they produce sparse, coarse intensity distributions that differ structurally from the continuous photometric variations that pretrained vision models expect. Accumulated event images exhibit histogram clustering around only 10--20 distinct intensity levels per 256 bins, lacking the smooth gradations characteristic of natural textures. Spatially, many regions are binary: either active where brightness changes occurred, or inactive where no change was detected. This creates sharp on-off transitions rather than gradual shading, as visualized in Fig.~\ref{fig:domain_gap}. More critically, events encode relative changes rather than absolute appearance, necessitating downstream networks to infer photometric content from inherently \emph{incomplete} information. This mismatch can notably degrade performance when event maps are fed into networks pretrained on natural image statistics such as DINOv2~\citep{oquab2023dinov2}.

\begin{figure}[t]
    \centering
    \includegraphics[width=0.75\linewidth]{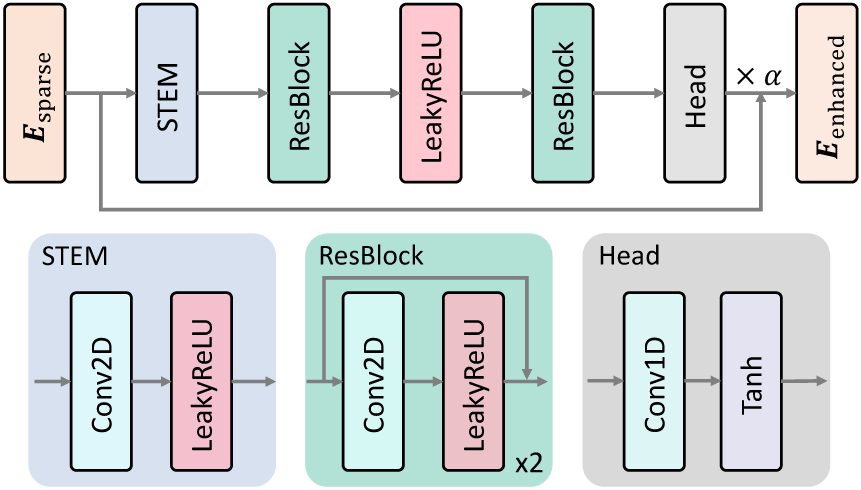}
    \vspace{-4pt}
    \caption{\textbf{Architecture of the Event Enhancement Module.} A lightweight residual network ($\sim$15K params.) with the learnable weight $\alpha$ transforms sparse event intensities into feature-rich representations compatible with pretrained encoders.}
    \label{fig:enhancement}
    \vspace{-12pt}
\end{figure}

To bridge this domain gap, we introduce a lightweight neural module $f_{\theta}$ that transforms sparse event representations into feature-rich inputs compatible with pretrained encoders while preserving their geometric authenticity. The module employs a residual architecture (see Fig.~\ref{fig:enhancement}):
\begin{equation}
    \mathbf{I}_{\text{enh}} = \mathbf{I}_{\text{event}} + \alpha \cdot f_{\theta}(\mathbf{I}_{\text{event}}),
\end{equation}
where $\mathbf{I}_{\text{event}}$ is the normalized, accumulated event image, $\alpha$ is a learnable scalar (initialized to 0.1) controlling enhancement strength, and $f_{\theta}$ predicts an additive refinement. The residual connection ensures stability---even if $f_{\theta}$ produces suboptimal outputs, the original event data is preserved.

The module $f_{\theta}$ consists of a stem (3$\times$3 convolution expanding single-channel input to $C{=}16$ feature channels with LeakyReLU activation), a refinement stage (1-2 residual blocks maintaining $C$ channels, where each block sequentially applies 3$\times$3 convolution, LeakyReLU activation, another 3$\times$3 convolution, element-wise addition, and LeakyReLU activation to its input), and a head (1$\times$1 convolution projecting back to a single channel with Tanh activation). We intentionally keep the enhancement output single-channel because the input physical quantity is a grayscale event accumulation map; this minimizes unnecessary pseudo-color hallucination and keeps the module lightweight before duplicating channels only for compatibility with the frozen encoder. With only ~10K--15K parameters, this module adds minimal overhead ($<$5~ms per 512$\times$1,024 image on an RTX 4090 GPU), making it practical for real-time applications.

During training, $f_{\theta}$ learns to smooth sparse intensity clusters into continuous gradients, infer missing photometric details from spatial context (e.g., interpolating intensities in low-event regions), and suppress noise from spurious events. The learnable weight $\alpha$ typically converges to 0.3--0.5, indicating the enhancement contributes 30--50\% of the final signal while preserving original event structure.

Beyond pixel-wise reconstruction loss, we introduce a frequency-domain loss to explicitly recover high-frequency details absent in sparse event representations:
\begin{equation}
    \begin{split}
        \mathcal{L}_{\text{freq}} =& \|\,|\mathcal{F}(\mathbf{I}_p)| - |\; \mathcal{F}(\mathbf{I}_t)|\,\|_1 
        + \lambda_{\phi} \mathcal{L}_{\text{phase}} \label{eq:freq_loss} ,\\
        \mathcal{L}_{\text{phase}} =& \frac{1}{|\Omega|} \sum_{(u,v) \in \Omega} 
        \min\Big(\big|\angle\mathcal{F}(\mathbf{I}_p)_{uv} - \angle\mathcal{F}(\mathbf{I}_t)_{uv}\big|, \\
        & \qquad\qquad 2\pi - \big|\angle\mathcal{F}(\mathbf{I}_p)_{uv} - \angle\mathcal{F}(\mathbf{I}_t)_{uv}\big|\Big),    
    \end{split}
\end{equation}
where $\mathbf{I}_p$ and $\mathbf{I}_t$ denote the predicted and target panoramas, respectively, $\mathcal{F}$ denotes 2D FFT operation, $|\cdot|$ extracts magnitude spectrum, $\angle$ extracts phase, and $\Omega = \{(u,v) : |\mathcal{F}(\mathbf{I}_t)|_{uv} > \tau \cdot \text{mean}(|\mathcal{F}(\mathbf{I}_t)|)\}$ ($\tau{=}0.1$) filters low-magnitude frequencies to prevent noise amplification. The magnitude term enforces energy distribution matching, while the phase term (w/ wrapping handling via $\min$) preserves spatial structure.

We note that this loss complements spatial losses ($\mathcal{L}_1$, gradient) by ensuring spectral consistency, which correlates strongly with perceptual quality~\citep{zhang2018unreasonable}.

\subsection{Learned Spherical Reconstruction (LSR)}
\label{sec:reconstruction}

Standard convolutional networks often fail on spherical panoramas due to equirectangular distortions and periodic boundaries. Instead, we adopt a Vision Transformer (ViT) architecture~\citep{dosovitskiy2020image}, whose permutation-invariant self-attention naturally handles non-Euclidean geometry. Critically, we incorporate explicit 3D spatial embeddings derived from the preserved spherical coordinates $\mathbf{V}$ (Sec.~\ref{sec:spherical_mapping}), ensuring geometric consistency throughout the pipeline.

\paragraph{Architecture overview.}
The LSR architecture consists of three components. First, the enhanced event image $\mathbf{I}_{\text{enh}} \in \mathbb{R}^{H \times W}$ is duplicated to 3 channels (converting from grayscale to RGB) and passed through a frozen DINO-ViT-Large encoder~\citep{oquab2023dinov2}:
\begin{equation}
    \mathbf{F} = \text{DINO}\left(\text{repeat}(\mathbf{I}_{\text{enh}}, 3)\right).
\end{equation}
Here DINO provides semantically rich features learned from natural images, which help infer photometric content from sparse event data.
Second, unlike standard ViTs that use 2D sinusoidal embeddings, we leverage the 3D unit vectors $\mathbf{V}(u, v) = (v_x, v_y, v_z)$ for spherical position encoding:
\begin{equation}
    \mathbf{E}_{\text{pos}}(u, v) = \text{MLP}_{\text{pos}}\big(\mathbf{V}(u, v)\big),
\end{equation}
where $\text{MLP}_{\text{pos}}$ is a 3-layer MLP projecting 3D coordinates to 1,024-dimensional embeddings. This approach achieves geometric consistency by using the same coordinates from the mapping stage, ensuring that reconstruction respects the original camera trajectory. Moreover, 3D embeddings do not encounter equirectangular distortions inherent in $(u, v)$ pixel coordinates.
Finally, we employ a 12-layer Transformer decoder with \emph{2D-3D-aware Cross-Attention} to DINO features:
\vspace{-6pt}
\begin{align}
    \begin{split}
        \mathbf{Q} &= \mathbf{E}_{\text{pos}} + \mathbf{E}_\mathrm{embed},\\%\text{Embed}_{\text{learnable}}, \\
        \mathbf{K}, \mathbf{V} &= \text{Linear}(\mathbf{F}), \\
        \mathbf{H}_{\ell} &= \text{TransformerBlock}_{\ell}(\mathbf{Q}, \mathbf{K}, \mathbf{V}),
    \end{split}
\end{align}
where $\mathbf{E}_\mathrm{embed}$ is a learnable embedding.
 The final layer outputs per-pixel luma predictions via a 1$\times$1 convolution. Since the decoder operates on image patches/tokens rather than raw per-pixel full attention, the dominant attention cost depends on the token count induced by the projected spherical map resolution rather than an explicit $O((HW)^2)$ dense pixel-wise attention.

\paragraph{Training strategy.}
We jointly optimize the event enhancement module ($f_{\theta}$) and the reconstruction network ($g_{\phi}$) in an end-to-end manner using a combined loss:
\begin{equation}
    \mathcal{L}_{\text{total}} = \lambda_1 \mathcal{L}_{\text{L1}} + \lambda_e \mathcal{L}_{\text{edge}} 
    + \lambda_f \mathcal{L}_{\text{freq}},
    \label{eq:losses}
\end{equation}
where $\mathcal{L}_{\text{L1}} = \|\mathbf{I}_{\text{pred}} - \mathbf{I}_{\text{gt}}\|_1$ measures pixel-wise reconstruction error, $\mathcal{L}_{\text{edge}}$ penalizes gradient differences computed via Sobel filters to preserve edge sharpness, and $\mathcal{L}_{\text{freq}}$ is the frequency-domain loss defined in Sec.~\ref{sec:enhancement}. We set loss weights to $\lambda_1{=}1.0$, $\lambda_e{=}0.5$, $\lambda_f{=}0.3$.

For optimization, we employ cosine annealing with warm restarts~\citep{loshchilov2016sgdr}. After a 3-epoch linear warmup from $5{\times}10^{-6}$ to $5{\times}10^{-5}$, the learning rate cycles every 10 epochs between peak ($5{\times}10^{-5}$) and minimum ($5{\times}10^{-6}$). This schedule prevents premature convergence and enables progressive refinement across multiple learning phases. Training uses the AdamW optimizer ($\beta_1{=}0.9$, $\beta_2{=}0.999$, weight decay $10^{-2}$) with batch size 8 on an RTX 4090 GPU. We employ FP16 mixed precision and gradient clipping (max-norm 0.5) for numerical stability. During training, the Event Enhancement Module's residual weight $\alpha$ typically converges to 0.3--0.5, indicating the learned enhancement contributes 30--50\% of the final signal while preserving the original event structure.

\begin{figure}[t]
    \centering
    \includegraphics[width=\linewidth]{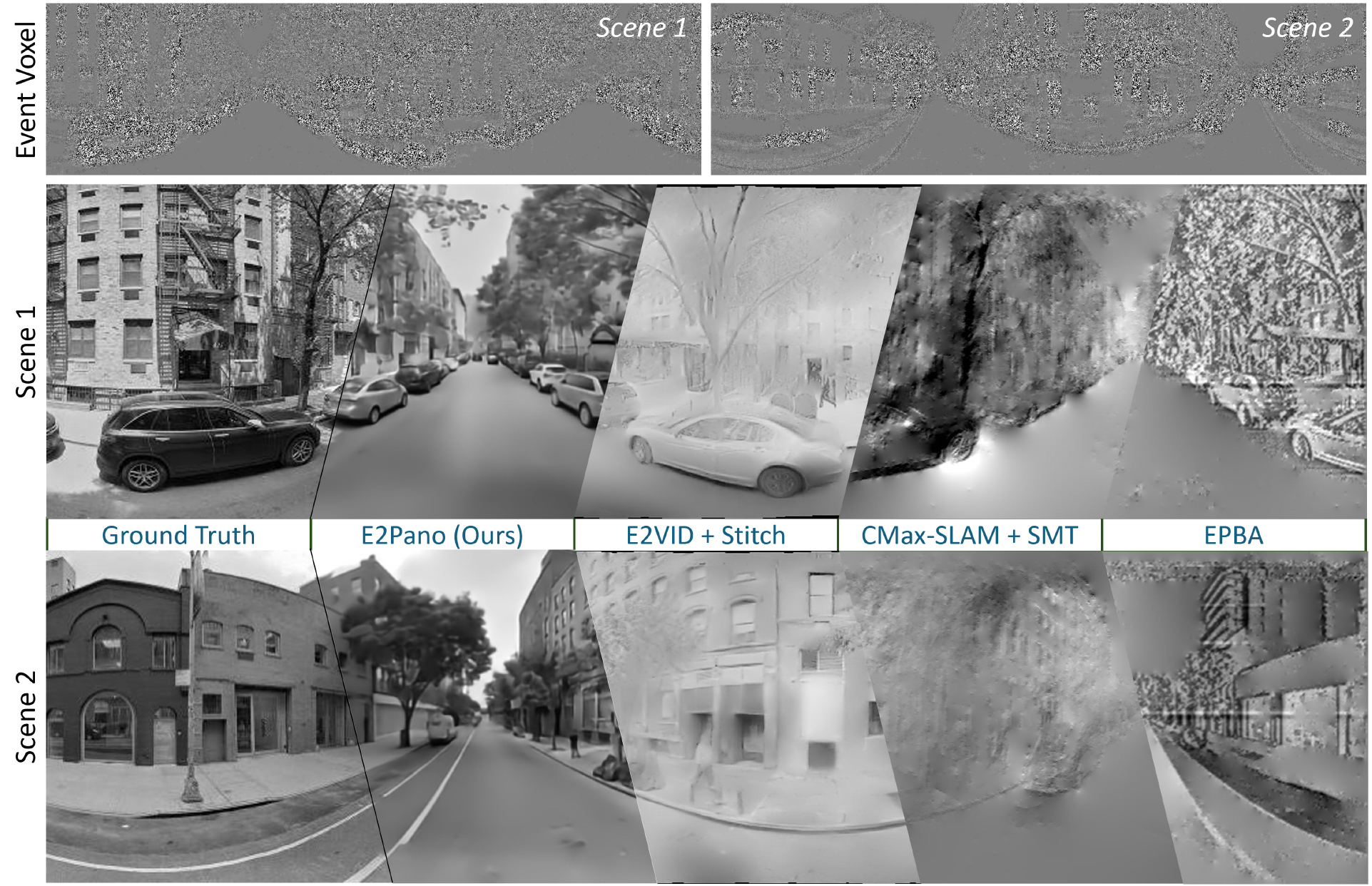}
    \vspace{-3pt}
    \caption{\textbf{Qualitative Results on \dataset{} Synthetic Test Set.} For each scene, \emph{left to right}: Ground truth panorama (Col.~1), E2Pano (Ours, Col.~2), E2VID~\citep{rebecq2019events} (E2VID++) + Stitch (Col.~3), CMax-SLAM~\citep{guo2024cmax} + SMT~\citep{smt} (Col.~4), and EPBA~\citep{guo2025event} (Col.~5). Our method yields sharper and more globally consistent reconstructions across the $360^{\circ}$ panoramic FoV. Event inputs are shown (\emph{top}).}
    \label{fig:synthetic_comparison}
    \vspace{-6pt}
\end{figure}

\begin{figure}[t]
    \centering
    \includegraphics[width=\linewidth]{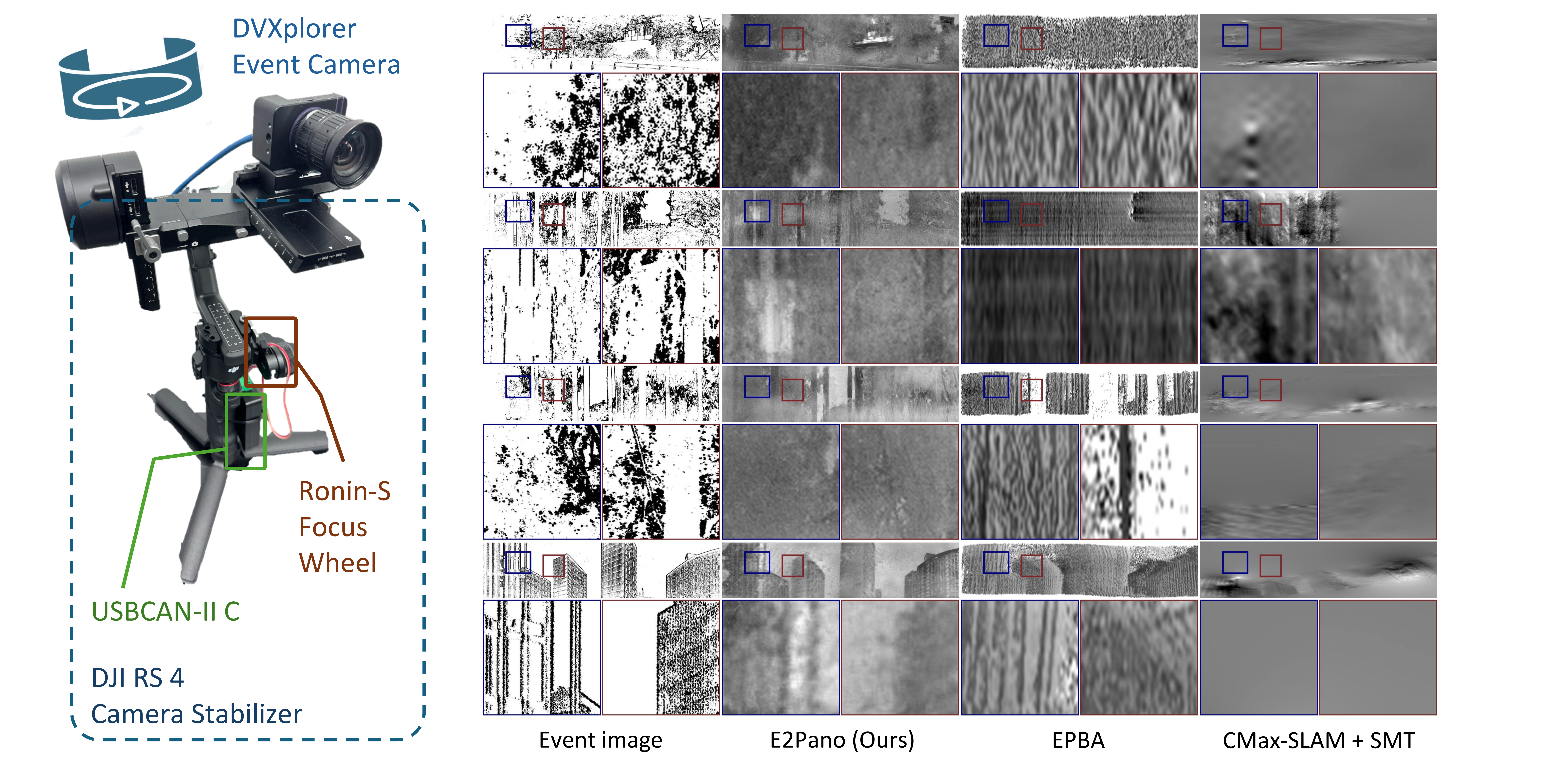}
    % \vspace{-6pt}
    \caption{\textbf{Real-World Acquisition Setup and Representative Captured-Scene Results.}
        \emph{Left-most}: Our event-based panoramic acquisition system comprising a DVXplorer camera and a DJI RS 4 stabilizer, enabling practical scanning acquisition in the wild. \emph{Right}: Representative qualitative results on a subset of real captured scenes. For each scene, the compared full pipelines are arranged \emph{left-to-right} as Event image, E2Pano (Ours), EPBA~\citep{guo2025event}, and CMax-SLAM~\citep{guo2024cmax}+SMT~\citep{smt}. For readability, each panorama is shown as a cropped view from the larger reconstruction. For each reconstructed panorama, two 3$\times$ zoom-in patches, marked by \textcolor{blue}{blue} and \textcolor{red}{dark red} boxes, are provided to highlight local detail preservation and reconstruction fidelity. Additional real-captured comparisons are provided in the \textbf{Supplementary Material}.
    }
    \label{fig:hardware_results}
    % \vspace{-3pt}
\end{figure}

\begin{figure}[t]
    \centering
    \includegraphics[width=\linewidth]{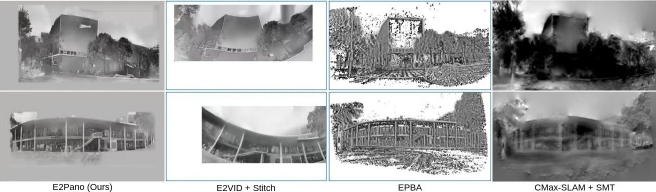}
    % \vspace{-15pt}
    \caption{\textbf{Qualitative Comparison on the Public EROAM-Captured Benchmark.}  For each scene, the compared full pipelines are arranged \emph{left-to-right} as E2Pano (Ours), E2VID~\citep{rebecq2019events} (retrained, E2VID++-style)+Stitch, EPBA~\citep{guo2025event}, and CMax-SLAM~\citep{guo2024cmax}+SMT~\citep{smt}. For readability, each panorama visualization is cropped from a larger reconstruction. Gray regions in our results indicate areas masked out due to insufficient valid observations after alignment or extremely sparse event support. Note that the E2VID-based stitching pipeline often fails to produce undistorted results because the intermediate perspective reconstructions are not optimized for spherical panorama generation.}
    \label{fig:eroam}
    % \vspace{-3pt}
\end{figure}

\begin{figure}[t]
    \centering
    \includegraphics[width=\linewidth]{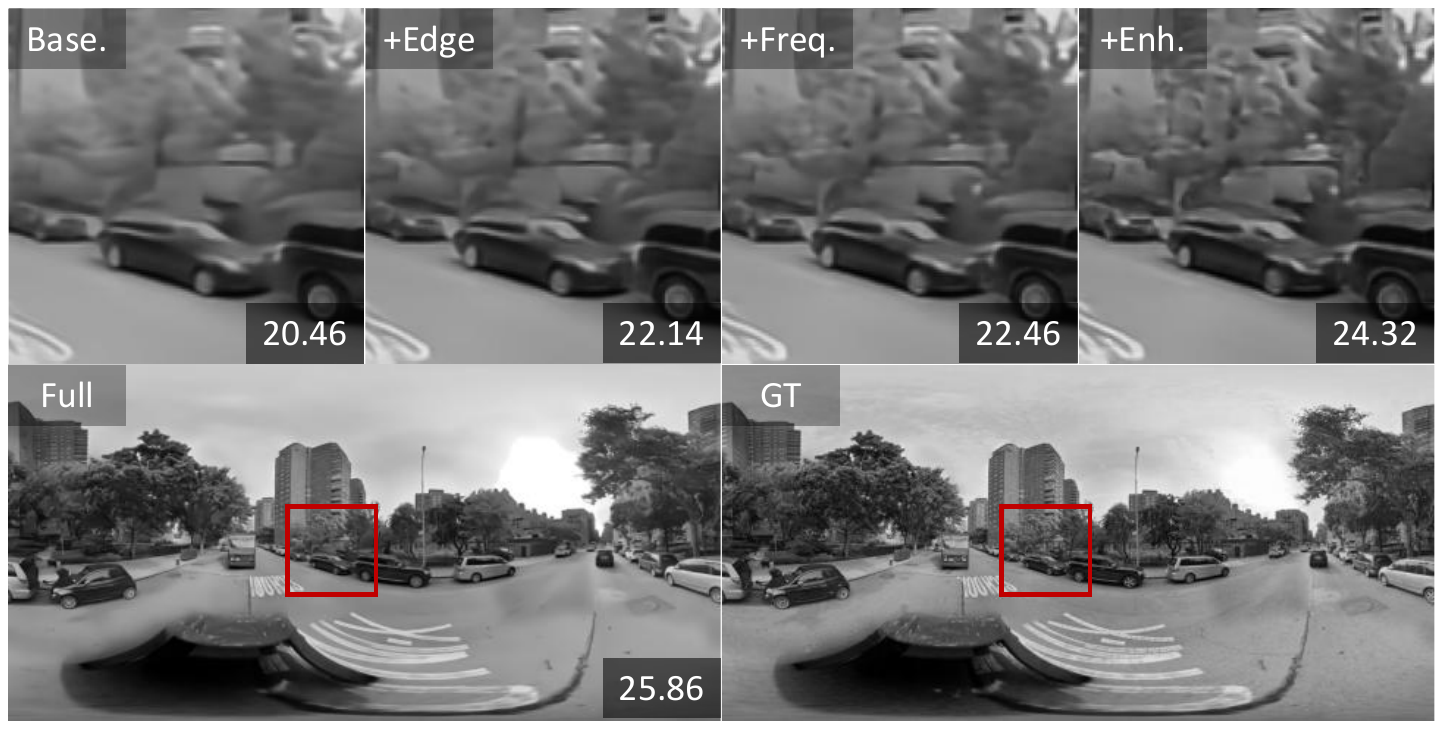}
    % \vspace{-15pt}
    \caption{\textbf{Ablation Study.} \emph{Top row, left to right}: Baseline, Baseline~+~$\mathcal{L}_{\mathrm{edge}}$, Baseline~+~$\mathcal{L}_{\mathrm{freq}}$, Baseline~+~Enhancement Module. \emph{Bottom row}: \emph{Full Model} that incorporates all components and the ground truth reference. Red squares denote the zoom-in region. 
    PSNR (dB) scores are shown in corners.}
    \label{fig:ablation_visual}
    % \vspace{-3pt}
\end{figure}

\section{Experiments}
\label{sec:experiments}

\subsection{Implementation Details}
\label{subsec:implement}

\emph{\textbf{\dataset{} dataset.}} Our constructed dataset comprises 4,370 synthetic scenes from OmniCity~\citep{li2023omnicity} and 30 real-world captures (10 indoors and 20 outdoors), are named \textit{\dataset{}-Sync} and \textit{\dataset{}-Real} respectively. Fifty synthetic scenes are held out from training as the test set. For synthetic scenes, we render perspective views from equirectangular panoramas under camera-center-fixed rotational motion and convert them to events using v2e~\citep{hu2021v2e}, producing $512\times1,\!024$ spherical grids paired with ground-truth luma panoramas. The full synthetic data-generation pipeline, including trajectory sampling, rendering details, and simulator settings, is described in the \textbf{Supplementary Material}.

For real-world captures, we use a custom scanning rig (Fig.~\ref{fig:hardware_results}, \emph{left-most}) consisting of a Prophesee DVXplorer event camera and a DJI RS 4 gimbal stabilizer controlled via Python SDK. The acquisition protocol is camera-center-fixed and rotation-dominated, typically covering about $360^{\circ} \times 50^{\circ}$ FoV at 10--30$^{\circ}$/s. Most captures follow yaw-dominant horizontal sweeps, while a smaller subset uses structured yaw-pitch scans; both are well approximated as pure rotation. The 30 captured sequences span 20 outdoor and 10 indoor scenes under daytime, dusk, and night lighting. Detailed hardware specifications, scan duration statistics, and scene lists are provided in the \textbf{Supplementary Material}. While diverse in appearance and lighting, this captured set still reflects a fixed acquisition rig and a limited family of rotational trajectories.

\vspace{3pt}
\noindent\emph{\textbf{Captured-data coverage.}} \textit{\dataset{}-Real} is designed to stress-test transfer across illumination and scene complexity rather than to maximize sequence count alone. It includes bright daytime outdoor scenes, mixed dusk transitions, and sparse night illumination, together with both highly textured structures and texture-poor regions such as walls and sky. Exact lighting ranges, event-rate statistics, depth variation, and per-scene metadata are summarized in the Supplementary Material.

\vspace{3pt}
\noindent\emph{\textbf{Training.}} We train exclusively on synthetic data
using the AdamW~\citep{loshchilovdecoupled} optimizer ($\beta_1 = 0.9$, $\beta_2 = 0.999$), batch size 8, and cosine annealing with warm restarts~\citep{loshchilov2016sgdr}. We apply a 3-epoch linear warmup from $5{\times}10^{-6}$ to $5{\times}10^{-5}$, followed by 10-epoch cosine cycles between $5{\times}10^{-5}$ and $5{\times}10^{-6}$. Loss weights in Eq.~\ref{eq:losses} are $\lambda_{\text{L1}}=1.0$, $\lambda_{\text{grad}}=0.5$, and $\lambda_{\text{freq}}=0.3$. All trainings are conducted on an NVIDIA RTX 4090 GPU.

\vspace{3pt}
\noindent\emph{\textbf{Baselines.}} We compare against three representative classes of baselines: (1) \emph{E2VID++}~\citep{rebecq2019events} + Stitch~\citep{brown2007automatic}, where our implementation follows the retrained threshold-robust E2VID setting; (2) \emph{CMax-SLAM}~\citep{guo2024cmax} + SMT~\citep{smt}, which represents rotational SLAM and event mosaicing pipelines; and (3) \emph{EPBA}~\citep{guo2025event}, which performs photometric bundle adjustment directly on events. These are complete-pipeline comparisons under identical input event streams. In all cases, the same scanning video / event stream is used as input, but each baseline estimates its own trajectory and reconstructs its own scene representation. Therefore, the final outputs reflect complete-system performance rather than a shared-front-end photometric back-end-only comparison.

\vspace{3pt}
\noindent\emph{\textbf{Metrics.}} For synthetic data with available ground truth, we report PSNR, SSIM, and LPIPS on reconstructed luma images. For real-world captures, where pixel-aligned ground truth is unavailable, the following reference-free metrics are used: Jensen-Shannon divergence (JS Div.)~\citep{lin2002divergence} to measure consistency between the gradient distributions of the reconstruction and event data; BRISQUE~\citep{mittal2012no} for assessing perceptual quality; and Event Alignment (E-Align)—that is, gradient-event correlation, detailed in the Supplementary Material—to evaluate geometric consistency. Processing time is measured end-to-end (from event loading to final panorama) on the same GPU for all methods.

\subsection{Benchmark on \dataset{}-Sync}
\label{subsec:synthetic_results}

We begin with the controlled synthetic benchmark on \dataset{}, where paired panorama ground truth is available. This setting provides the primary photometric evaluation of the proposed method, since reconstruction quality can be measured directly using PSNR, SSIM, and LPIPS.

Table~\ref{tab:synthetic_results} provides a quantitative comparison on the \dataset{} test set. Our method achieves substantially better scores (about 13~dB in PSNR$\uparrow$, 0.55 in SSIM$\uparrow$, and 0.018 in LPIPS$\downarrow$) than the evaluated stitching-based and optimization-based pipelines.
These results are supported by Figure~\ref{fig:synthetic_comparison}, where reconstructed panoramas for two different scenes are visualized. E2Pano (Col. 2) reconstructs sharper textures with better spherical continuity. E2VID+Stitch (Col. 3) often exhibits seam artifacts because it relies on perspective reconstructions followed by stitching rather than a spherical reconstruction model. CMax-SLAM+SMT (Col. 4) and EPBA (Col. 5) provide optimization-based full pipelines, but their final photometric quality remains affected by geometric instability and long-horizon iterative reconstruction on sparse event evidence. We use ES-ICP for geometric initialization and evaluate the performance of the complete pipeline in all comparisons.

\begin{table}[t]
    \centering
    \caption{Quantitative results on \dataset{}'s synthetic set (50 scenes). Metrics: PSNR, SSIM, and LPIPS.}
    \label{tab:synthetic_results}
    % \vspace{-6pt}
    \footnotesize
    \begin{adjustbox}{width=0.8\linewidth}
    \begin{tabular}{l|ccc}
    \toprule
    Method & PSNR (dB) $\uparrow$ & SSIM $\uparrow$ & LPIPS $\downarrow$ \\
    \midrule
    E2VID~\citep{rebecq2019events} (E2VID++) + Stitch & 10.41 & 0.007 & 0.303 \\
    CMax-SLAM~\citep{guo2024cmax} + SMT~\citep{smt} & 12.24 & 0.224 & 0.651 \\
    EPBA~\citep{guo2025event} & 11.34 & 0.079 & 0.496 \\
    \midrule
    Ours & \textbf{26.06} & \textbf{0.781} & \textbf{0.285} \\
    \bottomrule
    \end{tabular}
    \end{adjustbox}
    % \vspace{-9pt}
\end{table}

\begin{figure}[t]
    \centering
    \includegraphics[width=\linewidth]{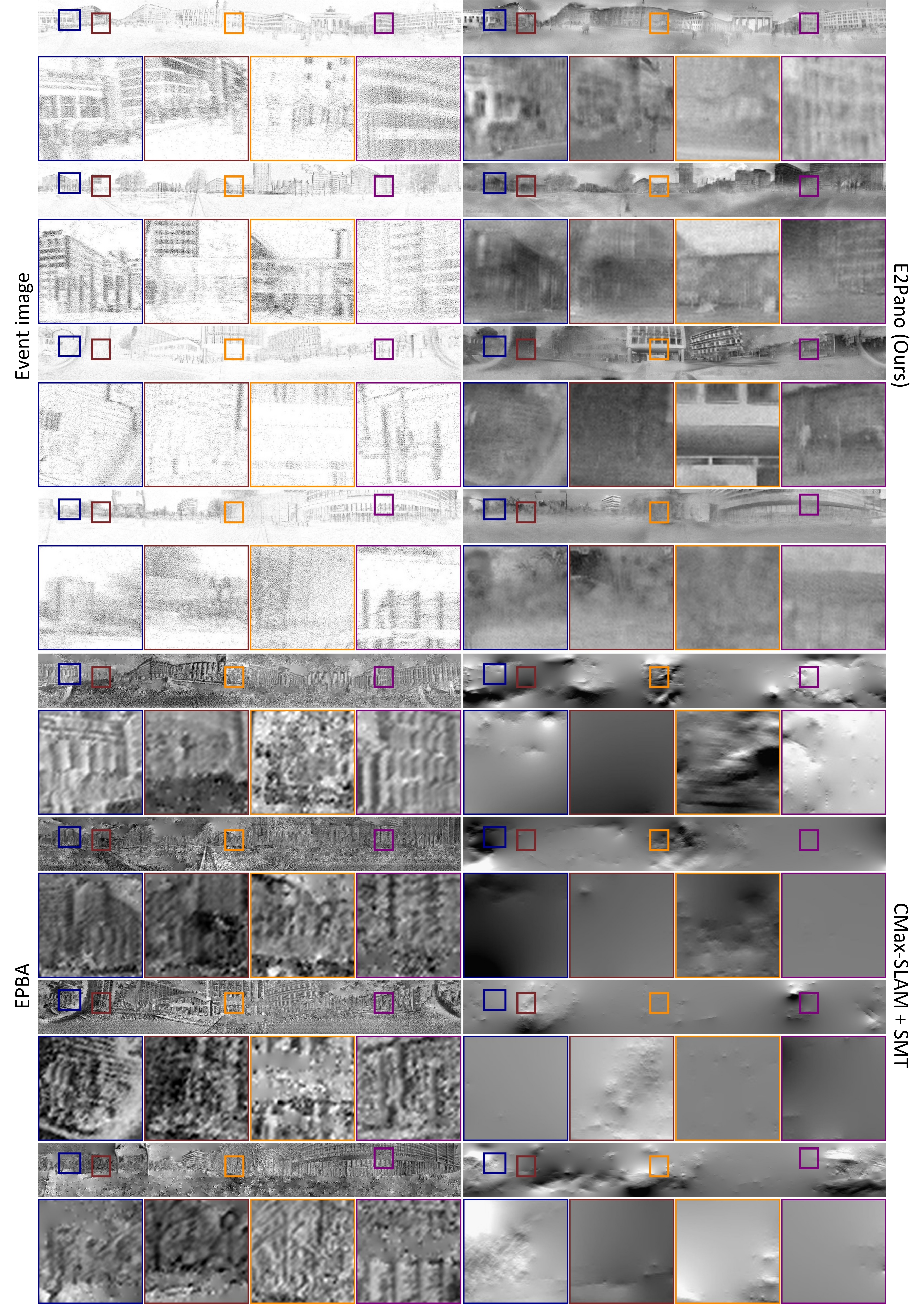}
    \caption{\textbf{Qualitative Comparison on the ECRot Dataset.} Columns from left to right show the event image, E2Pano (Ours), EPBA~\citep{guo2025event}, and CMax-SLAM~\citep{guo2024cmax}. This comparison evaluates visual transfer beyond \dataset{}-Sync on public real-captured event data under an adapted rotational-panorama protocol. For each reconstructed panorama, four zoom-in patches (marked with \textcolor{blue}{blue}, \textcolor{red}{dark red}, \textcolor{orange}{orange}, and \textcolor{purple}{purple} square boxes) at 3$\times$ magnification are presented, highlighting fine details and reconstruction fidelity.}
    \label{fig:ecrot}
\end{figure}

\subsection{Evaluation on Public Real-Captured Benchmarks}
\label{subsec:public_real}

We next evaluate transfer beyond the proposed \dataset{} on public real-captured event datasets: ECRot and EROAM-captured (\cref{fig:ecrot}, \cref{fig:eroam}). This section is intended to provide cross-dataset evidence using data that was not collected with our own rig. Because publicly available event datasets are not fully matched to our rotational panoramic setting, we adapt them to the closest compatible evaluation protocol and present qualitative comparisons under this adapted setup. 

Qualitative evidence on these public real-captured sequences is valuable because it tests transfer beyond \dataset{} without relying on our own acquisition rig. Since the available public data are only partially aligned with the rotational panoramic setting studied here, we use this section to present visual comparison rather than to define the primary quantitative benchmark.

\subsection{Qualitative Assessment on \dataset{}-Real}
\label{subsec:captured_qualitative}

Finally, we present qualitative comparisons on \dataset{}-Real to demonstrate practical deployment at larger scale and higher output resolution. Since these scenes are captured with our own rig and do not provide pixel-aligned ground truth, we use them as application-style visual evidence of how the full system behaves in realistic scans.

\paragraph{Qualitative analysis.}
Figure~\ref{fig:hardware_results} shows that our method consistently produces cleaner and more coherent panoramas than the competing full pipelines on our captured scenes. The most visible gains appear in thin structures, high-contrast boundaries, and repeated architectural details, where E2Pano preserves sharper edges and more stable local textures while reducing the heavy noise, geometric inconsistency, or over-smoothed appearance observed in the baselines. In particular, optimization-based baselines depend on trajectory estimation and scene reconstruction being jointly refined, so pose error and alignment drift can propagate into the final panorama and make the recovered appearance deviate from the real scene. By contrast, our learning-based photometric stage avoids such complex explicit camera-pose optimization in the reconstruction stage and therefore often yields results that are more visually consistent with the underlying scene content. We still observe failure cases in regions with insufficient event support (e.g., textureless walls) or local scene motion, where holes or incorrect coverage may appear. Additional scene-level comparisons and reference-free analyzes are provided in the \textbf{Supplementary Material}.

\paragraph{Failure evidence and boundary cases.}
Representative failure visualizations are provided in the \textbf{Supplementary Material}. In particular, the geometric front-end is most vulnerable to four situations: very high-speed rotations that reduce overlap between successive observations, multi-row or non-ideal scan patterns that deviate from the pure-rotation assumption, repetitive textures that create ambiguous correspondences, and texture-less sky or wall regions that provide too few constraints for stable registration. These failures are not merely presentation issues; they directly define the operating boundary of the current pipeline because photometric reconstruction depends on successful spherical alignment.

\subsection{Ablation Studies}
\label{subsec:ablation}

We first ablate each component individually in Table~\ref{tab:ablation} by adding it to the baseline. Edge loss ($\mathcal{L}_{\text{edge}}$) alone provides 1.68~dB PSNR gain, validating its role in preserving boundary structures. Frequency loss ($\mathcal{L}_{\text{freq}}$) contributes 2.00~dB improvement, demonstrating the importance of high-frequency detail recovery. The Enhancement Module yields the largest individual gain of 3.86~dB, confirming its critical role in bridging the event-image domain gap. 

The full model, integrating all components with joint optimization, achieves 5.40~dB improvement over the baseline, demonstrating that these components complement each other effectively. The edge loss produces the most direct structural benefit because it constrains spatial discontinuities explicitly, while the frequency loss acts as a complementary regularizer that improves texture continuity and spectral consistency even when its standalone PSNR gain is smaller than that of the Enhancement Module. As visualized in Figure~\ref{fig:ablation_visual}, the baseline produces noisy outputs, while the full model recovers sharp edges, fine textures, and smoother intensity gradations.

\begin{table}[t]
    \centering
    \caption{Ablation study of each component. Results are calculated on the validation set (10\% in total) of \dataset{}.}
    \label{tab:ablation}
    \footnotesize
    \begin{adjustbox}{width=0.7\linewidth}
    \begin{tabular}{l|ccc}
    \toprule
    Configuration & PSNR (dB) $\uparrow$ & SSIM $\uparrow$ & LPIPS $\downarrow$ \\
    \midrule
    Baseline & 20.46 & 0.6800 & 0.408 \\
    \quad + Edge Loss & 22.14 & 0.7084 & 0.372 \\
    \quad + Frequency Loss & 22.46 & 0.7194 & 0.354 \\
    \quad + Enhancement Module & 24.32 & 0.7517 & 0.322 \\
    Full Model & \textbf{25.86} & \textbf{0.7826} & \textbf{0.300} \\
    \bottomrule
    \end{tabular}
    \end{adjustbox}
    % \vspace{-6pt}
\end{table}

\subsection{Timing and Efficiency Analysis}
\label{subsec:timing}

Table~\ref{tab:timing} compares processing time on a representative scene (250K events, $512\times1,\!024$ output resolution). Our method processes the full pipeline (spherical mapping + enhancement + reconstruction) in 0.20~seconds (200~ms). For CMax-SLAM+SMT (5.7~sec) and EPBA (30~sec), the reported numbers correspond only to their event-alignment stages. Therefore, these timings are not directly comparable as uniform end-to-end measurements; instead, they are intended to show that the learned photometric stage in E2Pano adds very low overhead once a spherical event map has been formed. Note that the Event Enhancement Module adds only a negligible computation overhead ($<$1~ms), validating its lightweight design.
Combined with faster scanning hardware solutions, our results suggest that low-latency panoramic reconstruction from rotational scans is feasible, although a fully standardized runtime comparison remains future work.

\begin{table}[t]
    \centering
    \caption{Processing time comparison (RTX~4090, 250K events, $512\times1,\!024$ output). For CMax-SLAM and EPBA, only event alignment is timed.}
    \label{tab:timing}
    \footnotesize
    \begin{adjustbox}{width=0.7\linewidth}
    \begin{tabular}{l|cc}
    \toprule
    Method & Time (s)$\downarrow$ & Speedup \\
    \midrule
    CMax-SLAM~\citep{guo2024cmax} & 5.7 & 1.0$\times$ \\
    EPBA~\citep{guo2025event} & 30.0 & 0.2$\times$ \\
    E2VID~\citep{rebecq2019events} (E2VID++) + Stitch & 6.0 & 0.9$\times$ \\
    \midrule
    Ours (Full Pipeline) & \textbf{0.20} & \textbf{28.0$\times$} \\
    \bottomrule
    \end{tabular}
    \end{adjustbox}
    % \vspace{-8pt}
\end{table}

\section{Conclusion and Discussion}
\label{sec:conclusion}
In this work, we present \emph{E2Pano}, a geometry-guided event-to-panorama pipeline with an end-to-end learnable photometric reconstruction stage under rotational scanning.
Unlike prior optimization-based methods with prohibitive costs or learning-based approaches constrained to narrow-FoV planar projections, our geometry-aware design reduces photometric reconstruction cost while preserving spherical consistency for this acquisition setting.
We also construct \emph{PanoScan}, a comprehensive dataset comprising 4,370 synthetic and 30 real-world scenes, as a training set and evaluation benchmark for the event-to-panorama reconstruction task.

By preserving real spherical coordinates throughout the entire pipeline, E2Pano effectively bridges the event-image domain gap via a lightweight Enhancement Module and frequency-domain supervision.
Furthermore, a spherical Transformer with 3D positional embeddings is employed for high-fidelity photometric reconstruction.
Experimental results on PanoScan demonstrate improved reconstruction quality and encouraging transfer to real captures under our acquisition protocol, paving the way for event-based VR/AR solutions, virtual tours, and immersive panoramic media.

\vspace{3pt}
\emph{\textbf{Limitations \& Future Work.}}
Our current pipeline depends on the quality of ES-ICP-based rotational alignment; when the geometric front-end fails, the photometric stage can inherit holes, misalignment, or incomplete coverage. As shown by the supplementary failure cases, the most challenging scenarios include very high-speed rotations, multi-row scans that deviate from the pure-rotation assumption, repetitive textures that create ambiguous matches, and texture-less sky or wall regions with insufficient event support. The method is currently limited to grayscale luma reconstruction, and the real-world evaluation set, while diverse in scene type and lighting, is still collected with a fixed rig and a limited family of rotational trajectories. In addition, the current formulation targets near-pure rotational scanning rather than general 6-DoF motion, and texture-sparse or event-sparse regions can still lead to ambiguous reconstructions.
Future work includes stronger cross-dataset evaluation on public rotational benchmarks, stronger reconstructed-video baselines, color panorama reconstruction under sparse event triggers~\citep{scheerlinck2019ced}, self-supervised adaptation to reduce dependence on paired synthetic data, and more robust geometric front-ends for challenging scans and dynamic scenes.

\clearpage
\bibliographystyle{iclr2027_conference}
\bibliography{references}

% \appendix
% \input{sections/appendix.tex}
\end{document}